\documentclass{article}

    \PassOptionsToPackage{numbers, compress}{natbib}

    \usepackage[preprint]{neurips_2025}

\usepackage[utf8]{inputenc} 
\usepackage[T1]{fontenc}    
\usepackage{hyperref}       
\usepackage{url}            
\usepackage{booktabs}       
\usepackage{amsfonts}       
\usepackage{nicefrac}       
\usepackage{microtype}      
\usepackage{amsmath}
\usepackage{amssymb}
\usepackage{mathtools}
\usepackage{amsthm}
\usepackage{multirow}
\usepackage[table]{xcolor}
\usepackage{graphicx}
\usepackage{adjustbox}
\usepackage{array}
\usepackage{booktabs}
\usepackage{dsfont}
\usepackage{makecell}
\usepackage{wrapfig}
\usepackage{algorithmic}
\usepackage{algorithm}
\usepackage{enumitem}

\title{Distribution-Conditional Generation: \\ From Class Distribution to Creative Generation}

\author{%
  Fu Feng\textsuperscript{\rm 1,2}\quad Yucheng Xie\textsuperscript{\rm 1,2}\quad Xu Yang\textsuperscript{\rm 1,2}\quad Jing Wang\textsuperscript{\rm 1,2}\thanks{Corresponding authors}\quad Xin Geng\textsuperscript{\rm 1,2}\footnotemark[1] \\
  School of Computer Science and Engineering, Southeast University, Nanjing, China\\
  Key Laboratory of New Generation Artificial Intelligence Technology and Its Interdisciplinary \\Applications (Southeast University), Ministry of Education, China \\
  \texttt{\{fufeng, xieyc, xuyang\_palm, wangjing91, xgeng\}@seu.edu.cn} \\
}

\begin{document}

\maketitle

\begin{abstract}
    Text-to-image (T2I) diffusion models are effective at producing semantically aligned images, but their reliance on training data distributions limits their ability to synthesize truly novel, out-of-distribution concepts. Existing methods typically enhance creativity by combining pairs of known concepts, yielding compositions that, while out-of-distribution, remain linguistically describable and bounded within the existing semantic space.
    Inspired by the soft probabilistic outputs of classifiers on ambiguous inputs, we propose Distribution-Conditional Generation, a novel formulation that models creativity as image synthesis conditioned on class distributions, enabling semantically unconstrained creative generation.
    Building on this, we propose DisTok, an encoder-decoder framework that maps class distributions into a latent space and decodes them into tokens of creative concept.
    DisTok maintains a dynamic concept pool and iteratively sampling and fusing concept pairs, enabling the generation of tokens aligned with increasingly complex class distributions.
    To enforce distributional consistency, latent vectors sampled from a Gaussian prior are decoded into tokens and rendered into images, whose class distributions—predicted by a vision-language model—supervise the alignment between input distributions and the visual semantics of generated tokens. The resulting tokens are added to the concept pool for subsequent composition.
    Extensive experiments demonstrate that DisTok, by unifying distribution-conditioned fusion and sampling-based synthesis, enables efficient and flexible token-level generation, achieving state-of-the-art performance with superior text-image alignment and human preference scores.
\end{abstract}

\section{Introduction}
\begin{figure*}[t]
  \centering
  \includegraphics[width=\linewidth]{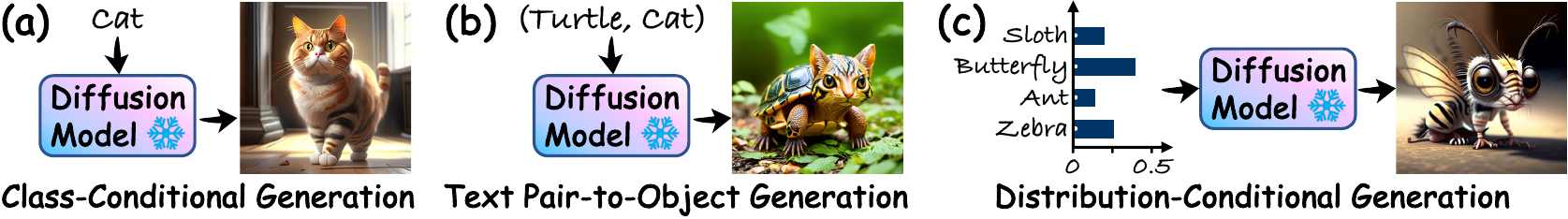}
  \vspace{-0.25in}
  \caption{(a) Traditional class-conditional generation maps a single label to a concept. (b) Text pair-to-object generation fuses two known concepts into a novel one. (c) In contrast, our proposed distribution-conditional generation accepts arbitrary semantic distributions over multiple classes, enabling controllable, fine-grained creativity beyond simple two-token interpolation.}
  \label{fig:moti}
  \vspace{-0.2in}
\end{figure*}

Recent advances in text-to-image (T2I) models, such as Stable Diffusion 3~\cite{esser2024scaling} and Midjourney~\cite{midjourney}, have showcased impressive capabilities in generating photorealistic and semantically aligned images from natural language prompts~\cite{wu2024multimodal, lin2024evaluating, chen2024tailored}, exemplified by class-conditional generation task in Fig.~\ref{fig:moti}a, highlighting the maturity of generative models in reproducing known visual concepts.

However, the strong alignment exhibited by these models largely stems from their reliance on training data distributions, which constrains their ability to generalize beyond seen concepts and limits out-of-distribution creativity. 
As a result, recent efforts have focused on fostering creativity in generative models~\cite{richardson2024conceptlab, litp2o, feng2024redefining, han2025enhancing}, aiming to enable the synthesis of novel and meaningful content beyond replication—particularly under minimal or ambiguous linguistic supervision.

BASS~\cite{litp2o} defines creativity as the combinatorial synthesis of known concept pairs (e.g., \texttt{(Turtle, Cat)}) under the Text Pair-to-Object (TP2O) task (Fig.~\ref{fig:moti}b). 
While enabling out-of-domain generation, the resulting images remain semantically grounded and easily interpretable, limiting their departure from established conceptual boundaries.
ConceptLab~\cite{richardson2024conceptlab} introduces the Creative Text-to-Image (CT2I) task, defining creativity as the generation of novel, linguistically inexpressible concepts by intentionally departing from known semantic regions. However, this unbounded exploration excludes human-guided priors entirely, making it difficult to generate meaningful yet novel concepts such as ``a creative creature that resembles a cat'', complicating both controllability and evaluation.

A fundamental challenge in defining creativity lies in specifying \textbf{\textit{what constitutes a novel concept}}. 
Classification models offer a useful perspective: when confronted with ambiguous or out-of-distribution inputs, they typically output soft probability distributions over known classes, capturing partial affinities without assigning a definitive label. 
This suggests that novelty can emerge from blending familiar concepts in varying proportions.
Inspired by this, we invert the classification process by generating images conditioned on class distributions—a formulation we term \textbf{Distribution-Conditional Generation} (Fig.\ref{fig:moti}c).
By conditioning on class distributions rather than discrete prompts or fixed concept pairs, Distribution-Conditional Generation facilitates exploration of the creative concept space while retaining interpretability and controllability.

Existing methods for combinatorial creativity typically rely on reference images or tokens. 
BASS~\cite{litp2o} employs heuristic sampling over candidate images to identify creative compositions, while CreTok~\cite{feng2024redefining} leverages a universal token to encode combinatorial semantics.
Personalization-based approaches~\cite{liew2022magicmix, zhang2024diffmorpher, xiong2024novel} generate blended outputs via semantic interpolation within the diffusion process.
While effective for binary concept fusion, existing methods offer limited controllability and cannot accommodate class distributions involving more than two concepts. To overcome this limitation, we introduce \textbf{DisTok}, an encoder-decoder framework that maps arbitrary class \underline{dis}tributions to creative concept \underline{tok}ens via a learnable \textbf{Distribution Encoder} and \textbf{Creative Decoder}.

DisTok maintains a Concept Pool $\mathcal{P}$ initialized with tokens of known concept and jointly optimizes two complementary objectives to generate \textbf{\textit{increasingly complex}} and \textbf{\textit{semantically consistent}} creativity. 
To facilitate an increasingly complex semantic composition, DisTok continuously encode concept pairs $(c_1, c_2)$ sampled from $\mathcal{P}$ and decodes them into a single concept token $t_{\text{crt}}$. This process is indirectly supervised by aligning an adaptive prompt (``a photo of a $\langle t_{\text{crt}}\rangle$'') with restrictive prompts (``a $c_1$ $c_2$'' and ``a photo of a cute pet'').
To enhance semantic consistency, DisTok periodically samples latent vectors from a Gaussian prior, decodes them into novel tokens, and synthesizes corresponding images. A pretrained vision-language model is then queried to infer class distributions (e.g., via ``What animal is in the photo?''). 
The predicted distribution is used to supervise the alignment between the input class distribution and the visual semantics of the output token. Resulting tokens are added to $\mathcal{P}$, enabling subsequent semantic composition toward increasingly complex class distributions.

DisTok is trained on common classes from \textit{CangJie}\cite{feng2024redefining} and, once trained, can \textbf{\textit{directly generate tokens of creative concept}} without additional updates or optimization, enabling efficient inference. 
Compared to state-of-the-art (SOTA) creative generation methods, DisTok achieves a 13$\times$ speedup over BASS (3s vs. 40s) by eliminating candidate sampling, and a 40$\times$ acceleration over ConceptLab\cite{richardson2024conceptlab} (3s vs. 120s) through single-pass latent sampling, eliminating costly iterative optimization.
The tokens produced by DisTok exhibit strong semantic consistency across prompts, and the resulting images outperform SOTA diffusion models in both human preference and text-image alignment.
Additional evaluations with GPT-4o~\cite{achiam2023gpt} and user studies further validate DisTok’s superiority in concept integration and originality, underscoring its effectiveness for creative image synthesis.

Our contributions are summarized as follows:
(1) We introduce Distribution-Conditional Generation, a novel formulation that models creativity as conditional generation from class distributions, enabling fine-grained, controllable composition beyond discrete concept pairs.
(2) We propose DisTok, an encoder-decoder framework that maps class distributions to a latent space and decodes them into tokens of creative concepts.
DisTok unifies distribution-conditioned fusion and sampling-based synthesis, enabling efficient and flexible token-level generation.
(3) Extensive experiments validate DisTok’s state-of-the-art performance in creative generation, consistently outperforming both standard text-to-image models and creativity-focused methods in efficiency, diversity, and semantic alignment, as confirmed by human evaluations and quantitative metrics.

\section{Related Work}
\textbf{Creative Generation.}
Text-to-image (T2I) models have demonstrated impressive capabilities in synthesizing semantically coherent images~\cite{mou2024t2i, yue2024few, hu2024one}, yet extending these models toward creative generation remains a central challenge for machine intelligence~\cite{mateja2021towards, mazzone2019art, feng2023genes}. Current approaches predominantly rely on combinatorial creativity, generating novel concepts by merging or modifying known ones. For instance, BASS~\cite{litp2o} employs sampling strategies to combine pairs of distinct concepts, CreTok~\cite{feng2024redefining} introduces a dedicated token explicitly encoding combinational semantics, and ConceptLab~\cite{richardson2024conceptlab} promotes novelty by progressively decreasing similarity to existing concepts. 
Beyond concept-level combination, diffusion-based methods such as ProCreate~\cite{lu2024procreate} and C3~\cite{han2025enhancing} enhance creativity via latent-space interventions—repelling outputs from reference images or amplifying latent features. 
However, these techniques remain constrained by discrete pairwise combinations or indirect latent manipulations, limiting semantic flexibility and controllability. To overcome these limitations, we propose \textit{distribution-conditional generation}, where creative generation is directly conditioned on semantic distributions over known classes, thereby enabling more fine-grained, controllable, and personalized creative outputs beyond discrete textual prompts or concept pairs.

\textbf{Personalized Visual Content Generation.}
Personalization aims to generate coherent representations of target concepts from a few exemplar images~\cite{peng2024portraitbooth, ruiz2024hyperdreambooth}.
Early methods such as Textual Inversion~\cite{gal2023an, ruiz2023dreambooth} embed new identities into unique tokens via optimization on reference images.
Recent work emphasizes modularity and compositionality, with PartCraft~\cite{ng2024partcraft} and Chirpy3D~\cite{ng2025chirpy3d} modeling visual entities at the part level to enable flexible recombination.
Beyond token-level adaptation, some approaches adapt specific network components~\cite{han2023svdiff, zhang2024spectrum, kumari2023multi} or introduce external adapters~\cite{ye2023ip, chen2024artadapter, dorfman2025ip} to improve flexibility.
Other methods, such as MagicMix~\cite{liew2022magicmix}, DiffMorpher~\cite{zhang2024diffmorpher}, and ATIH~\cite{xiong2024novel}, promote innovative visual representations by interpolating semantic representations during the diffusion process~\cite{zhong2024multi}.
In contrast, DisTok decodes conceptual mixtures directly from class distributions, producing tokens that implicitly capture abstract concepts, and enabling zero-shot generation of unseen concepts without additional training or adaptation.

\section{Methods}

\subsection{Overview of DisTok}
\begin{figure*}[t]
  \centering
  \includegraphics[width=\linewidth]{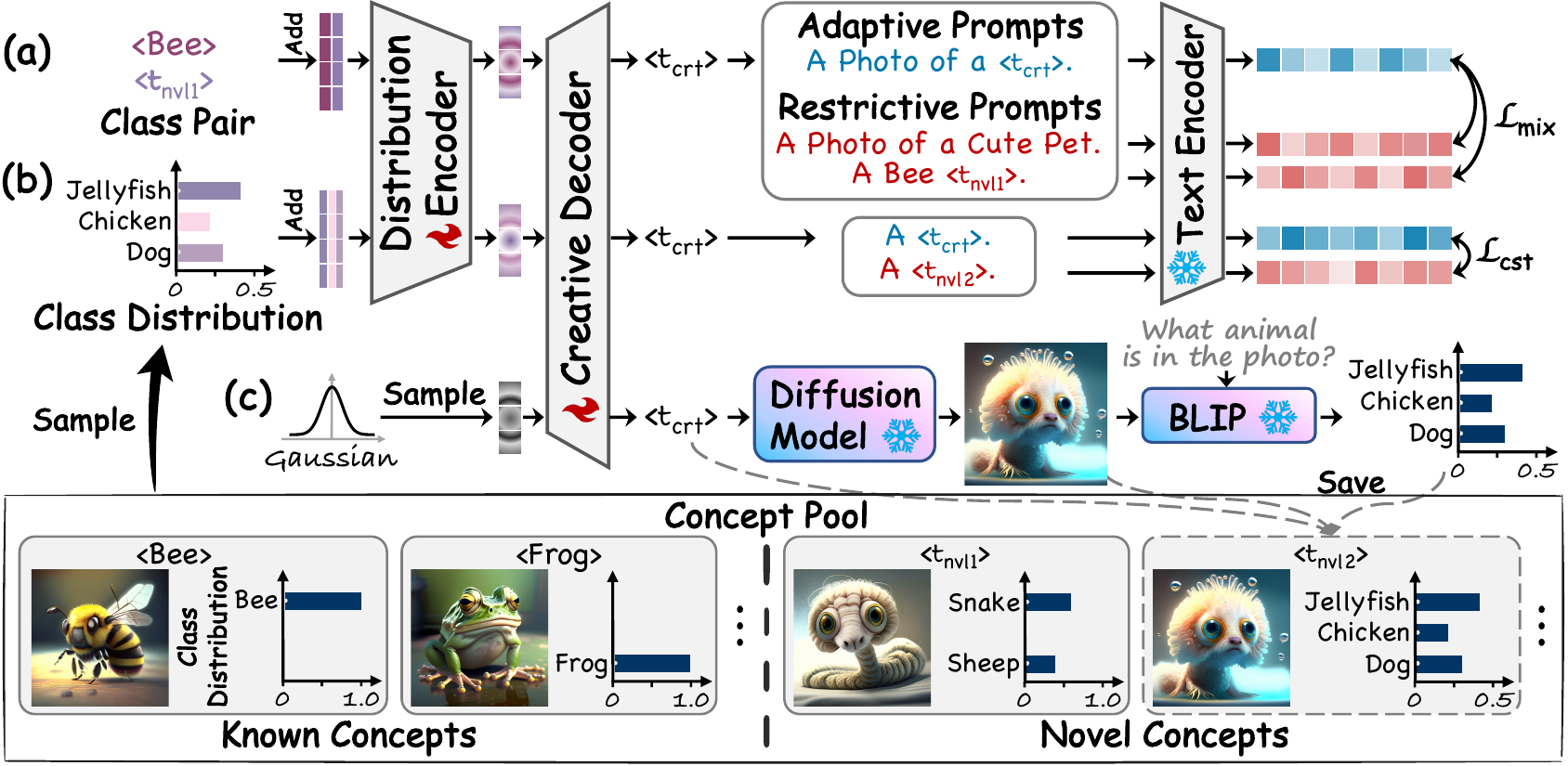}
  \vspace{-0.25in}
  \caption{Overview of DisTok. At each training step, DisTok performs either \textbf{(a)} concept combination by sampling a class pair to train the decoder to generate tokens aligned with increasingly complex class distributions, or \textbf{(b)} distribution consistency supervision by sampling a class distribution to align the encoder and decoder with the visual semantics of generated tokens. 
  \textbf{(c)} Latent vectors are periodically sampled and decoded into tokens, with class distributions predicted by a vision-language model. Resulting tokens and distributions are saved for subsequent combination and supervision.}
  \label{fig:main}
  \vspace{-0.14in}
\end{figure*}

DisTok generates tokens of creative concepts conditioned on class distributions, enabling the modeling of complex, ambiguous, and hard-to-describe visual concepts.
Unlike existing token-based approaches to combinatorial creativity~\cite{feng2024redefining, richardson2024conceptlab, litp2o}, DisTok adopts an encoder-decoder architecture, where the \textbf{\textit{Distribution Encoder}} $\mathcal{E}_{\text{dis}}$ projects class distributions into a latent space, and the \textbf{\textit{Creative Decoder}} $\mathcal{D}_{\text{tok}}$ maps latent representations into tokens that can be seamlessly integrated into prompts (Fig.~\ref{fig:main}). 
Formally, given a class distribution $p_c \in \Delta^{K}$, $\mathcal{E}_{\text{dis}}$ projects $p_c$ into a latent representation $z=\mathcal{E}_{\text{dis}}(p_c)\in \mathbb{R}^{\delta}$, which is then decoded into a creative concept token $t_{\text{crt}}=\mathcal{D}_{\text{tok}}(z)\in \mathbb{R}^d$ ($\delta \ll d$).

To progressively model more complex and semantically coherent concepts, DisTok maintains a \textbf{\textit{Concept Pool}} $\mathcal{P}$ initialized with tokens of known concept tokens and employs three key components to jointly train $\mathcal{E}_{\text{dis}}$ and $\mathcal{D}_{\text{tok}}$: 
\begin{itemize}[leftmargin=20pt,itemsep=0pt]
    \item \textbf{Continuous Concept Combination} (Sec.~\ref{sec:combine}). 
    DisTok continuously samples class pairs from $\mathcal{P}$ and trains to combine them into a single concept, progressively enabling $\mathcal{D}_{\text{tok}}$ to decode tokens corresponding to increasingly complex class distributions from latent representations.
    \item \textbf{Class Distribution Estimation} (Sec.~\ref{sec:estimate}). DisTok generates novel concepts by decoding randomly sampled latent vectors and predicting their class distributions via VLMs, with the resulting tokens and distributions added to $\mathcal{P}$ as novel concepts for further training. 
    \item \textbf{Class Distribution Consistency} (Sec.~\ref{sec:consist}). 
    DisTok enforces consistency between the input class distribution and the one predicted from generated images, providing direct supervision for both $\mathcal{E}_{\text{dis}}$ and $\mathcal{D}_{\text{tok}}$ using novel concepts from $\mathcal{P}$.
\end{itemize}

\subsection{Continuous Concept Combination via Class Pairs}
\label{sec:combine}
To train the Distribution Encoder $\mathcal{E}_{\text{dis}}$ and Creative Decoder $\mathcal{D}_{\text{tok}}$ for effective class distribution encoding and creative concept generation, it is first necessary to synthesize distribution-describable images without access to dedicated training samples.

BASS~\cite{litp2o} and CreTok~\cite{feng2024redefining} demonstrate that basic combinatorial creativity can be achieved through text-pair fusion, while ConceptLab~\cite{richardson2024conceptlab} shows that newly generated tokens can be recursively combined.
These observations suggest that continuously composing existing or novel concepts enables the construction of images aligned with specific class distributions.
For example, given classes $c_1, c_2, c_3\in \mathcal{C}$, a target class distribution $p_c=(\text{0.25}, \text{0.25}, \text{0.5})$ over ($c_1$, $c_2$, $c_3$) can be approximately synthesized by first combining ($c_1$, $c_2$) into an intermediate concept $c_{\text{inter}}$, then merging $c_{\text{inter}}$ with $c_3$.

Thus, DisTok iteratively samples class pairs from the current Concept Pool $\mathcal{P}$ and performs concept fusion to progressively enhance the Creative Decoder's ability to synthesize images corresponding to increasingly complex class distributions, as illustrated in Fig.~\ref{fig:main}a.
Specifically, given two concepts (i.e. classes) $c_1, c_2\in \mathcal{P}$, their corresponding tokens $t_1$, $t_2$ are combined and encoded into a latent vector $z=\mathcal{E}_{\text{dis}}(t_1 + t_2)$, which is subsequently decoded into a creative token $t_{\text{crt}}=\mathcal{D}_{\text{tok}}(z)$. 

To facilitate the fusion of $c_1$ and $c_2$, we construct an \textit{adaptive prompt} $q_a=\text{``a photo of a } \langle t_{\text{crt}}\rangle\text{''}$ and a \textit{restrictive prompt} $q_r(c_1, c_2) = \text{``a } c_1\ c_2\text{''}$, following the combination strategy of CreTok~\cite{feng2024redefining}.
To further align generation with human-preferred semantics, we introduce an auxiliary restrictive prompt $q_s = \text{``a photo of a cute pet''}$.
Semantic fusion is encouraged by minimizing:
\begin{equation}
    \tilde{\mathcal{L}}_{\text{mix}} = (1 - \text{cos}(E(q_r), E(q_a))) + (1 - \text{cos}(E(q_s), E(q_a))).
\label{eq:mix}
\end{equation}
where $\text{cos}(a, b) = \frac{a\cdot b}{||a||||b||}$ denotes cosine similarity, and $E(\cdot)$ is the text encoder (i.e., CLIP-L/14~\cite{radford2021learning} in Kandinsky 2.1).
To prevent overfitting to a dominant concept, which can artificially inflate similarity by disproportionately emphasizing one concept while neglecting others, we introduce thresholding to regulate concept integration.
The final loss is defined as:
\begin{equation}
    \mathcal{L}_{\text{mix}} = (1 - \min[\text{cos}(E(q_r), E(q_a)), \theta_1]) + (1 - \min[\text{cos}(E(q_s), E(q_a)), \theta_2]).
\label{eq:mix_the}
\end{equation}
where $\theta_1$ and $\theta_2$ are predefined thresholds controlling the maximum allowed similarity.

\subsection{Class Distribution Estimation via Vision-Language Models}
\label{sec:estimate}
Building on the continuous concept combination strategy introduced in Section~\ref{sec:combine}, DisTok progressively enhances the Creative Decoder’s ability to generate tokens $t_{\text{crt}}$ whose corresponding images $x_{\text{crt}}$ generated by diffusion models $x_{\text{crt}}=\mathcal{G}_{\text{diff}}(t_{\text{crt}})$ reflect increasingly complex class distributions.
However, this training provides only indirect supervision, as it only encourages semantic fusion by combining concept pairs $(c_1, c_2)$ while the mixing loss $\mathcal{L}_{\text{mix}}$ (Eq.~\eqref{eq:mix_the}) supervises only prompt-level semantics without explicitly regulating the contribution of each concept within $t_{\text{crt}}$.

To address this limitation, we further introduce a more precise supervision mechanism based on a pretrained VLM.
Given a generated image $x_{\text{crt}}$, we query the VLM with a question such as ``What animal is in the photo?'' and obtain its token-level output logits $p_{\text{vlm}}(c|x_{\text{crt}})$ over the known concepts $\mathcal{C}_{\text{knw}}\subset \mathcal{P}$.
The predicted class distribution, obtained by applying softmax to $p_{\text{vlm}}(c|x_{\text{crt}})$, serves as structured supervision for enforcing distributional consistency during training of $\mathcal{E}_{\text{dis}}$ and $\mathcal{D}_{\text{tok}}$ (Sec.~\ref{sec:consist}), as illustrated in Fig.~\ref{fig:main}c.

The images $x_{\text{crt}}$ for distribution prediction are generated by sampling latent vectors $z \sim \mathcal{N}(0, I)$ from a standard Gaussian distribution, decoding them into creative tokens $t_{\text{crt}} = \mathcal{D}_{\text{tok}}(z)$, and synthesizing the corresponding images $x_{\text{crt}}=\mathcal{G}_{\text{diff}}(t_{\text{crt}})$.
To ensure that random samples yield meaningful concepts rather than semantically irrelevant noise, we regularize the latent space by minimizing:
\begin{equation} 
    \mathcal{L}_{\text{reg}} = \frac{1}{\sigma(z)^2} \mathbb{E}_{z}\left[|\mu(z)|_2^2\right], 
\end{equation} 
where $\mu(z)$ and $\sigma(z)$ denote the mean and standard deviation of the latent vectors, respectively.

This regularization ensures that, during inference, new tokens and images can be generated by sampling $z$ from any distribution $\mathcal{Q}$ satisfying $\mathbb{E}_{z\sim \mathcal{Q}}[l]=0$ and $\text{Var}_{z\sim \mathcal{Q}}[l]=1$, enabling open-ended distributional creativity without requiring explicit reference concepts.

This process of sampling latent vectors for class distribution prediction is periodically performed during training. The resulting tokens and their associated distributions are added to $\mathcal{P}$ as novel concepts if they satisfy $\max_{c} p_{\text{vlm}}(c \mid x_{\text{crt}}) < \tau$, where $\tau$ is a predefined threshold controlling concept novelty.
These novel concepts enable higher-order semantic composition (Sec.~\ref{sec:combine}) and provide precious supervision for enforcing distributional consistency (Sec.~\ref{sec:consist}).

\subsection{Class Distribution Consistency via Input-Output Alignment}
\label{sec:consist}
Building upon the expanded $\mathcal{P}$, which includes novel concepts with more accurately estimated class distributions from the VLM, we further supervise the $\mathcal{E}_{dis}$ to improve distributional encoding and enforce consistency between input class distributions and the decoder outputs.

Specifically, during training, we sample a novel token $t_{\text{nvl}}$ along with its associated class distribution $p_c$.
We compute the weighted combination of relevant known tokens as $\Sigma_i p_c(i)t_i$, where $t_i$ denotes the token vector associated with class $c_i$.
This combined token is then encoded into a latent representation $z=\mathcal{E}_{\text{dis}}(\Sigma_i p_c(i)t_i)$, and subsequently decoded into a creative token $t_{\text{crt}}$.

To enforce distributional consistency, we minimize the cosine distance between the text embeddings of the generated token $t_{\text{crt}}$ and the sampled novel token $t_{\text{nvl}}$ from $\mathcal{P}$.
\begin{equation} 
    \mathcal{L}_{\text{cst}} = 1 - \text{cos}(E(t_{\text{crt}}), E(t_{\text{nvl}})), 
\end{equation} 
where $E(\cdot)$ denotes the text encoder and $\text{cos}(\cdot, \cdot)$ denotes the cosine similarity, both defined in Eq.~\eqref{eq:mix}.

Unlike the indirect supervision provided by recursive concept fusion (Eq.~\eqref{eq:mix_the}), VLM-based consistency explicitly grounds creative tokens in visual semantics.
By directly aligning generated tokens with their intended class distributions, as reflected in the synthesized images, it encourages $\mathcal{E}_{\text{dis}}$ to capture distributional semantics more faithfully, enabling DisTok to produce tokens of creative concepts that are both visually novel and semantically coherent with multi-concept compositions.

\subsection{Iterative Training of DisTok}
Each training iteration consists of $n$ sampling steps. At each step, we randomly select either the recursive concept fusion strategy (Sec.\ref{sec:combine}) or the distributional consistency objective (Sec.\ref{sec:consist}) to supervise the training of $\mathcal{E}_{\text{dis}}$ and $\mathcal{D}_{\text{tok}}$.
The overall training objective is defined as: 
\begin{equation} 
    \mathcal{L}_{\text{total}} = \frac{1}{n} \sum_{i=1}^n \left(\alpha \mathbb{I}_{\text{mix}}^{(i)} \mathcal{L}_{\text{mix}}^{(i)} + \beta \mathbb{I}_{\text{cst}}^{(i)} \mathcal{L}_{\text{cst}}^{(i)} + \gamma \mathcal{L}_{\text{reg}}^{(i)} \right), 
\end{equation} 
where $\mathbb{I}_{\text{mix}}^{(i)}, \mathbb{I}_{\text{cst}}^{(i)}\in \{0,1\}$ with $\mathbb{I}_{\text{mix}}^{(i)}+\mathbb{I}_{\text{cst}}^{(i)}=1$, and $\alpha$, $\beta$, $\gamma$ are weighting coefficient.

\section{Experiments}
\textbf{Datasets.}
We adopt the \textit{CangJie}~\cite{feng2024redefining} dataset, which contains 60 base concepts across animals and plants.
Originally designed for the TP2O task, it includes 30 text pairs for evaluating two-concept combinatorial creativity.
To support distribution-conditional generation, we extend \textit{CangJie} by constructing 30 class distributions by randomly combining and weighting the original concepts.

\textbf{Experimental Setup.}
Our implementation is based on Kandinsky 2.1~\cite{razzhigaev2023kandinsky}, which employs CLIP-L/14~\cite{radford2021learning} as the text encoder. 
The distribution encoder $\mathcal{E}_{\text{dis}}$ and creative decoder $\mathcal{D}_{\text{tok}}$ are two-layer MLPs with a hidden dimension of 768 and a latent space dimension of 20.
DisTok is trained for 20K steps on an NVIDIA 4090 GPU with a batch size of 1 and gradient accumulation over $n=8$ steps.
We use a cosine learning rate schedule with an initial rate of 0.0005, and set $\alpha=1$, $\beta=1$, $\gamma=0.001$ and $\tau=0.85$.


\textbf{Evaluation Metrics.}
We evaluate model performance using VQAScore~\cite{lin2025evaluating}, PickScore~\cite{kirstain2023pick}, and ImageReward~\cite{xu2024imagereward}, which assess alignment with text prompts, aesthetic quality, and human preferences.
In addition, we employ GPT-4o~\cite{openai2023gpt4} and conduct a user study to evaluate creativity in terms of conceptual integration, originality, and aesthetics.

\section{Results}
\subsection{Performance on Distribution-Conditional Generation Task}
\begin{figure*}[t]
  \centering
  \includegraphics[width=0.91\linewidth]{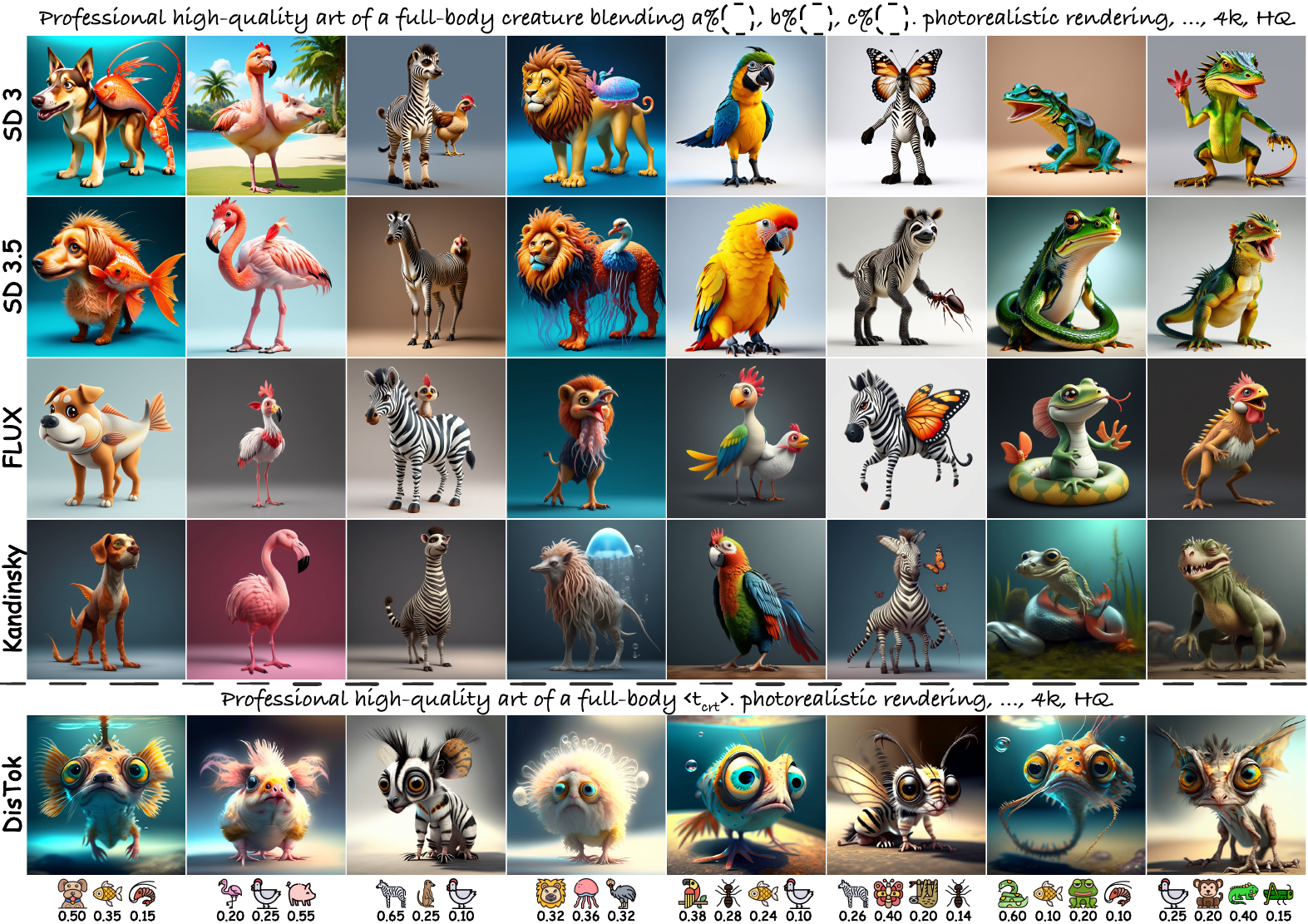}
  \vspace{-0.11in}
  \caption{Performance of DisTok on Distribution-Conditional Generation task. 
  }
  \label{fig:dcg}
  \vspace{-0.12in}
\end{figure*}

The Distribution-Conditional Generation task introduces a novel challenge in creative synthesis by requiring the generation of visual concepts conditioned on fine-grained semantic distributions over multiple known classes. 
We evaluate DisTok against state-of-the-art text-to-image (T2I) diffusion models, including Stable Diffusion 3~\cite{ramesh2022hierarchical}, Stable Diffusion 3.5~\cite{stability2024}, FLUX~\cite{black2024}, and Kandinsky~\cite{razzhigaev2023kandinsky}.

As shown in Fig.~\ref{fig:dcg}, while state-of-the-art diffusion models are effective at generating visually appealing images, they struggle with creative synthesis when conditioned on class distributions.
In the absence of explicit mechanisms for modeling such distributions, these models often fail to integrate three or more concepts coherently—leading to missing semantic elements, disjoint object representations, or visually fragmented compositions.
Moreover, they exhibit low sensitivity to the specified class proportions (e.g., ``55\% pig''), resulting in semantically unbalanced outputs and limited control over fine-grained concept composition.

In contrast, DisTok employs an encoder-decoder architecture that directly transforms class distributions into tokens of creative concepts (i.e., $\langle t_{\text{crt}\rangle}$), which are subsequently generated by diffusion models. 
This framework enables the generation of semantically coherent, visually integrated concepts that faithfully reflect the specified class distributions, supporting fine-grained and controllable synthesis beyond discrete prompt-based generation.

\subsection{Performance on Text Pair-to-Object Task}
\begin{figure*}[t]
  \centering
  \includegraphics[width=\linewidth]{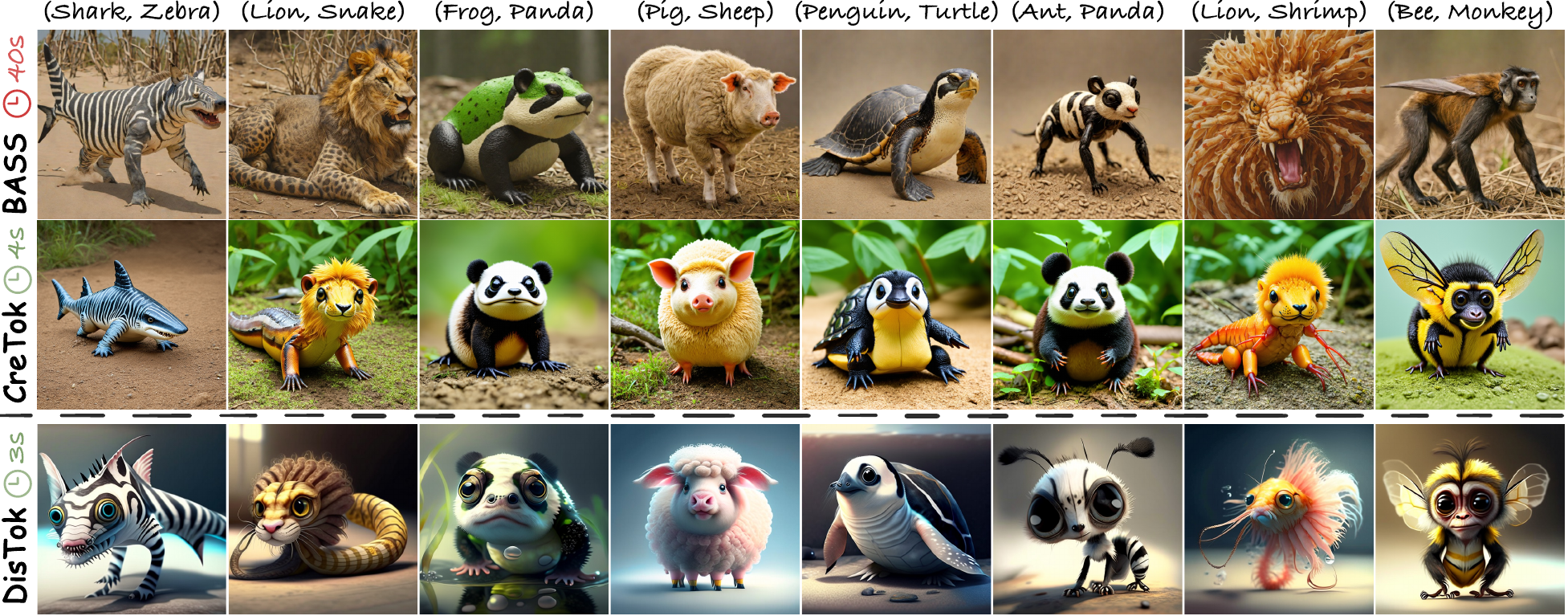}
  \vspace{-0.25in}
  \caption{Performance of DisTok on TP2O task.}
  \label{fig:tp2o}
  \vspace{-0.15in}
\end{figure*}

The Text Pair-to-Object (TP2O) task, which aims to fuse two known concepts into an integral representation, can be viewed as a special case of distribution-conditional generation with a uniform distribution over two classes. DisTok naturally generalizes to this setting and outperforms TP2O-specific state-of-the-art methods, including BASS~\cite{litp2o} and CreTok~\cite{feng2024redefining}.

As shown in Fig.~\ref{fig:tp2o}, DisTok generates visually coherent and semantically integrated hybrids that better reflect both input classes. 
Compared to BASS, which requires costly iterative sampling and candidate filtering ($\sim$40s per concept), DisTok directly encodes a concept pair and decodes a creative token in a single pass, incurring negligible additional overhead over standard diffusion inference ($\sim$3s).

While CreTok successfully captures the semantics of combinatorial creativity through a shared token, enabling efficiency and flexibility, its fixed fusion mechanism constrains output diversity. As shown by its near-identical outputs for \texttt{(Lion, Snake)} and \texttt{(Lion, Shrimp)}, CreTok struggles to distinguish between similar concept pairs, reflecting limitations inherent to prompt-level fusion. In contrast, DisTok generates distinct and semantically coherent hybrids for each pair, demonstrating stronger compositional sensitivity and greater expressiveness.
Thus, by modeling combinatorial creativity in latent space and aligning generated concepts with human-preferred aesthetics, DisTok enables efficient, diverse, and coherent creative generation.

\subsection{Unconditional Creative Generation without Reference Concepts}
\begin{figure*}[t]
  \centering
  \includegraphics[width=\linewidth]{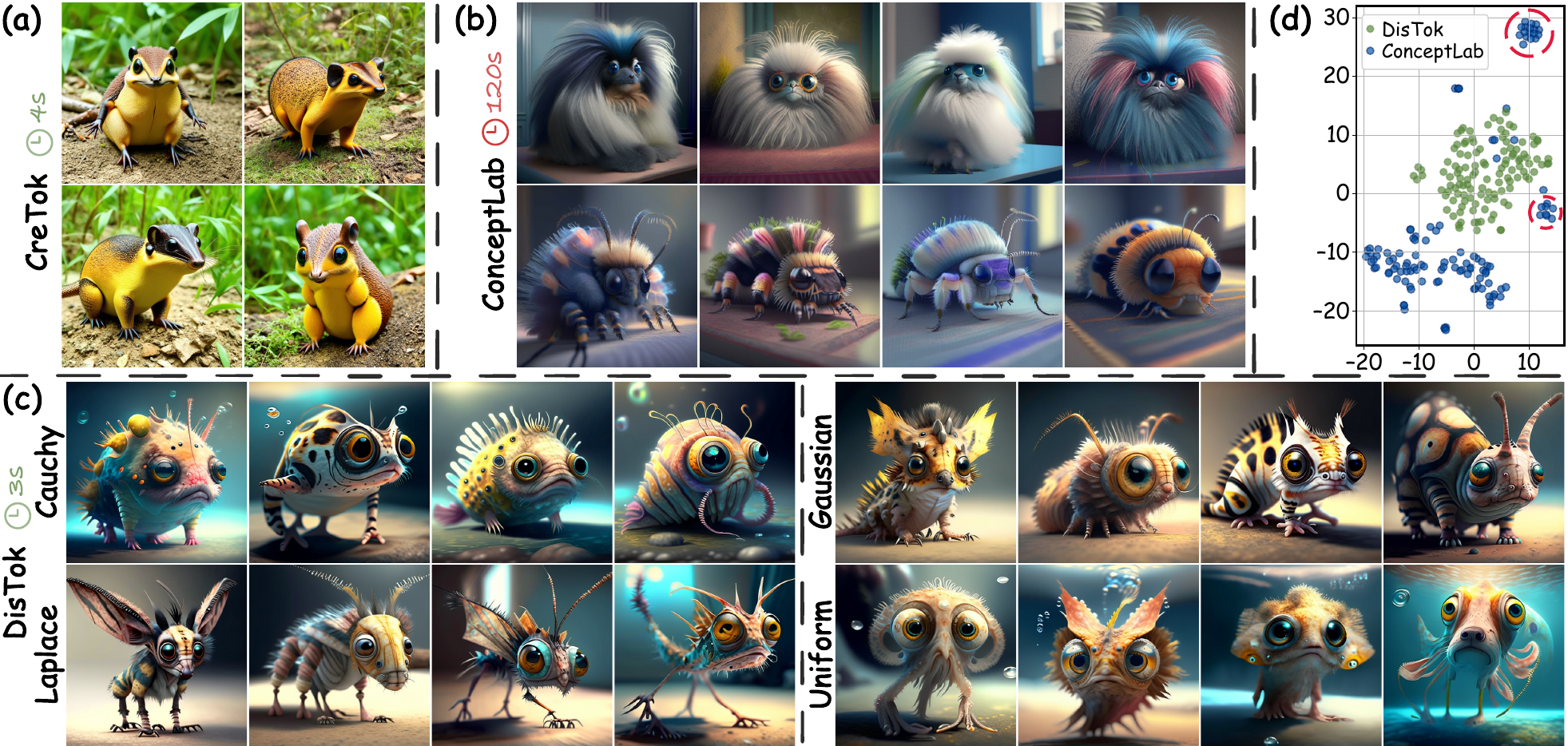}
  \vspace{-0.26in}
  \caption{Performance of DisTok in Direct Creative Generation without Reference Concepts.}
  \label{fig:conceptlab}
  \vspace{-0.1in}
\end{figure*}

DisTok provides a distinctive advantage over prior methods such as CreTok~\cite{feng2024redefining} and BASS~\cite{litp2o} by enabling reference-free creative generation. 
As shown in Fig.~\ref{fig:conceptlab}a, CreTok relies on explicit concept pairs and fails to produce diverse outputs without such guidance.
ConceptLab~\cite{richardson2024conceptlab}, enables unconstrained concept generation through iterative token optimization to deviate from known classes, but suffers from high computational cost ($\sim$120s per concept) due to its gradient-based search process.

In contrast, DisTok employs an encoder-decoder framework that enables direct sampling of latent vectors and decoding into creative tokens without reliance on external prompts.
This design enables efficient exploration of novel and diverse concepts beyond simple recombinations of existing ones, while eliminating iterative search and mitigating the limitations of prompt-dependent methods.

More importantly, ConceptLab~\cite{richardson2024conceptlab} remains semantically constrained, as each optimization step merely pushes tokens away from limited known classes, leading to concept clusters near sparse and repetitive boundaries (red circle in Fig.~\ref{fig:conceptlab}d). 
This limits diversity and novelty, particularly in large-scale generation.
In contrast, DisTok leverages a structured latent space regularized during training (Sec.~\ref{sec:estimate}), enabling \textbf{\textit{direct sampling}} from any zero-mean, unit-variance distribution (e.g., Gaussian, Laplace, Cauchy) without requiring iterative optimization. 
As shown in Fig.~\ref{fig:conceptlab}c, intra-distribution sampling yields substantial diversity, while inter-distribution sampling further enhances variability, demonstrating DisTok’s effectiveness as a unified framework for open-ended creative synthesis.

\subsection{Universality of DisTok in Generate Creativity with Diverse Styles}
\begin{figure*}[t]
  \centering
  \includegraphics[width=\linewidth]{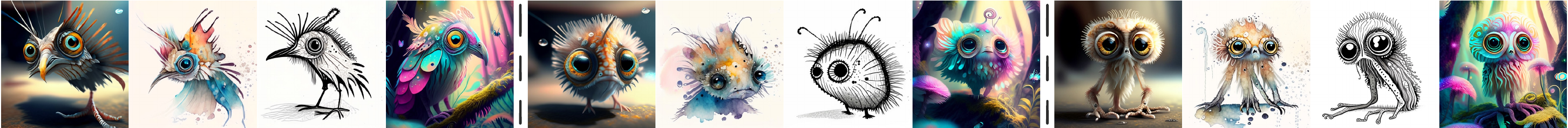}
  \vspace{-0.25in}
  \caption{Tokens generated by DisTok can be seamlessly combined with natural language prompts to support diverse styles while preserving conceptual consistency. See Appendix~\ref{app:style} for more styles.}
  \label{fig:style}
  \vspace{-0.14in}
\end{figure*}

DisTok demonstrates strong universality in style-aware creative generation by directly producing creative tokens that serve as reusable semantic anchors, enabling seamless style transfer via natural language prompts without retraining. As shown in Fig.~\ref{fig:style}, DisTok consistently preserves conceptual identity across diverse styles. In contrast, BASS~\cite{litp2o}, which samples and filters images without token representations, lacks the flexibility to adapt concepts through textual conditioning, while CreTok~\cite{feng2024redefining}, lacking explicit semantic anchoring, often fails to maintain coherence under stylistic shifts.

\subsection{Evaluation for Creativity}
\begin{wraptable}{r}{0.4\textwidth}
    \centering
    \setlength{\tabcolsep}{1 mm}
    \vspace{-0.25in}
    \caption{Quantitative Comparisons for Image-Text Alignment and Human Preference Ratings.}
    \resizebox{0.4\textwidth}{!}{
        \begin{tabular}{@{}lccc|c@{}}
        \toprule[1pt]
             & BASS & CreTok & \cellcolor{red!12}{DisTok} & \cellcolor{blue!12}{DisTok (DCG)} \\
             \midrule[0.5pt]
             VQAScore$\uparrow$~\cite{lin2025evaluating} & 0.667 & 0.695 & \cellcolor{red!12}{\textbf{0.840}} & \cellcolor{blue!12}{\textbf{0.734}}\\
             \midrule[0.5pt]
             PickScore$\uparrow$~\cite{kirstain2023pick} & 21.67 & 21.97 & \cellcolor{red!12}{\textbf{22.33}} & \cellcolor{blue!12}{\textbf{21.23}}\\
             ImageReward$\uparrow$~\cite{xu2024imagereward} & 0.387 & 1.018 & \cellcolor{red!12}{\textbf{1.168}} & \cellcolor{blue!12}{\textbf{0.661}}\\
             \bottomrule[1pt]
        \end{tabular}
        }
    \label{tab:human_pre}
    \vspace{-0.2in}
\end{wraptable}
\textbf{Quantitative Comparisons.} We evaluate DisTok on image-text alignment and on human preference under both text-pair and class-distribution conditions (Table~\ref{tab:human_pre}).
DisTok consistently outperforms BASS~\cite{litp2o} and CreTok~\cite{feng2024redefining} across all metrics on TP2O tasks, demonstrating its ability to synthesize semantically faithful and aesthetically preferred creative images. 
Although existing metrics may underestimate outputs with strong out-of-distribution characteristics, VQAScore still highlights DisTok’s strength in capturing fine-grained conditions (e.g., class distributions).

\renewcommand{\arraystretch}{0.8}
\begin{wraptable}{r}{0.4\textwidth}
    \centering
    \setlength{\tabcolsep}{1 mm}
    \vspace{-0.25in}
    \caption{Creativity evaluated by GPT-4o.}
    \resizebox{0.4 \textwidth}{!}{
        \begin{tabular}{@{}llccccc@{}}
        \toprule[1pt]
             & & Inte. & Align. & Orig. & Aesth. & \textit{Compr.} \\
             \midrule[0.5pt]
             \multirow{5}{*}{\rotatebox{90}{\small{\textbf{Distribution}}}} & SD 3~\cite{esser2024scaling} & 6.3$\pm$1.8 & 5.7$\pm$3.4 & 5.8$\pm$4.8 & 8.0$\pm$0.5 & \textit{6.5$\pm$2.0} \\
             & SD 3.5~\cite{stability2024} & 6.7$\pm$1.3 & 6.1$\pm$2.3 & 6.4$\pm$2.3 & 8.2$\pm$0.4 & \textit{6.9$\pm$1.2} \\
             & Kandin~\cite{razzhigaev2023kandinsky} & 7.7$\pm$1.5 & 7.4$\pm$2.6 & 7.3$\pm$2.2 & 8.6$\pm$0.4 & \textit{7.8$\pm$1.4} \\
             & FLUX~\cite{black2024} & 8.0$\pm$0.9 & 7.7$\pm$1.5 & 8.0$\pm$1.4 & 8.8$\pm$0.2 & \textit{8.1$\pm$0.9} \\
             & \cellcolor{blue!12}{DisTok} & \cellcolor{blue!12}{\textbf{9.2$\pm$0.2}} & \cellcolor{blue!12}{\textbf{9.2$\pm$0.2}} & \cellcolor{blue!12}{\textbf{9.8$\pm$0.1}} & \cellcolor{blue!12}{\textbf{9.9$\pm$0.1}} & \cellcolor{blue!12}{\textbf{\textit{9.5$\pm$0.1}}} \\
             \midrule
             \multirow{3}{*}{\rotatebox{90}{\small{\textbf{TP2O}}}} & BASS~\cite{litp2o} & 5.7$\pm$2.0 & 5.5$\pm$4.6 & 5.6$\pm$4.5 & 8.2$\pm$0.3 & \textit{6.3$\pm$2.1} \\
             & CreTok~\cite{feng2024redefining} & 7.3$\pm$2.3 & 7.7$\pm$5.3 & 7.1$\pm$3.0 & 8.8$\pm$0.2 & \textit{7.7$\pm$1.9} \\
             & \cellcolor{red!12}{DisTok} & \cellcolor{red!12}{\textbf{9.3$\pm$0.3}} & \cellcolor{red!12}{\textbf{9.9$\pm$0.1}} & \cellcolor{red!12}{\textbf{9.2$\pm$0.2}} & \cellcolor{red!12}{\textbf{9.1$\pm$0.3}} & \cellcolor{red!12}{\textbf{\textit{9.4$\pm$0.1}}} \\
             \bottomrule[1pt]
        \end{tabular}
    }
    \label{tab:gpt}
    \vspace{-0.2in}
\end{wraptable}
\textbf{Evaluation via GPT-4o.}
Given the limitations of existing metrics in capturing out-of-distribution creativity, we adopt GPT-4o for assessments (prompts in Appendix~\ref{app:prompt}).
Table~\ref{tab:gpt} shows that DisTok outperforms all SOTA diffusion models on distribution-conditional generation task, with notable gains in originality and aesthetics, highlighting its capacity to synthesize novel and creative concepts.

\textbf{User Study.}
We conduct a user study with 100 participants from diverse creative domains and higher education backgrounds (see Appendix~\ref{app:user} for details). Each participant evaluates 50 image pairs 
\begin{wraptable}{r}{0.5\textwidth}
    \centering
    \setlength{\tabcolsep}{0.8 mm}
    \vspace{-0.1in}
    \caption{Results of the user study.}
    \resizebox{0.5 \textwidth}{!}{
        \begin{tabular}{@{}lcccc|cc|c@{}}
        \toprule[1pt]
            & \multicolumn{4}{c|}{Distribution Conditional Generation} & \multicolumn{2}{c|}{TP2O} & Uncondition \\
            \cmidrule[0.5pt]{2-8}
             & SD 3 & SD 3.5 & Kandin & FLUX & BASS & CreTok & ConceptLab \\
             \midrule[0.5pt]
             DisTok & \textbf{312}:188 & \textbf{320}:180 & \textbf{355}:145 & \textbf{278}:222 & \textbf{396}:104 & \textbf{330}:170 & \textbf{308}:192 \\
             \bottomrule[1pt]
        \end{tabular}
        }
    \label{tab:user_study}
    \vspace{-0.15in}
\end{wraptable}
generated by DisTok and baseline methods across three tasks, selecting the image they find more creative. 
As shown in Table~\ref{tab:user_study}, DisTok consistently receives the majority of votes, outperforming both SOTA T2I models and creativity-focused baselines. These results highlight DisTok’s ability to synthesize semantically integrated, visually coherent, and creatively compelling concepts.

\subsection{Ablation and Analysis}
\begin{figure*}[t]
  \centering
  \includegraphics[width=\linewidth]{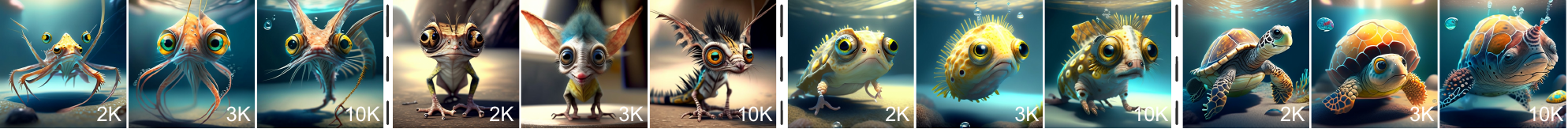}
  \vspace{-0.27in}
  \caption{Progressively complex creative concepts sampled at 2K, 3K, and 10K training steps.}
  \label{fig:complex}
  \vspace{-0.1in}
\end{figure*}
\textbf{Increasingly Complex Sementic through Concept Fusion.} 
DisTok decode creative concepts with increasingly complex class distributions by repeatedly sampling concept pairs from the Concept Pool and applying semantic fusion (Section~\ref{sec:combine}). To illustrate this progression, we visualize tokens decoded at different training stages. As shown in Fig.~\ref{fig:complex}, the resulting images exhibit a clear trajectory from simple compositions to semantically rich and entangled concepts, demonstrating DisTok’s capacity to model fine-grained compositional variation within its latent space.

\textbf{Distributional Consistency between Language and Visual Semantics.}
DisTok promotes alignment between input class distributions and the semantic content of generated images through distributional 
\begin{wraptable}{r}{0.2\textwidth}
    \centering
    \setlength{\tabcolsep}{0.8 mm}
    \vspace{-0.2in}
    \caption{Ablation of Distribution Consistency Supervision.}
    \resizebox{0.2 \textwidth}{!}{
        \begin{tabular}{@{}lcc@{}}
        \toprule[1pt]
             & w/o Cons. & DisTok \\
             \midrule[0.5pt]
             KL$\downarrow$ & 0.0732 & \textbf{0.0602} \\
             \bottomrule[1pt]
        \end{tabular}
        }
    \label{tab:abla_cons}
    \vspace{-0.14in}
\end{wraptable}
supervision (Section~\ref{sec:consist}). 
We evaluate its impact via an ablation study by removing this supervision and computing the KL divergence between the input token distribution and the visual distribution predicted by BLIP.
As shown in Table~\ref{tab:abla_cons}, removing this objective significantly increases divergence, highlighting the importance of distributional supervision in maintaining fine-grained semantic consistency during generation.

\textbf{Creativity Beyond Linguistic Expression.}
\begin{figure*}[t]
  \centering
  \includegraphics[width=\linewidth]{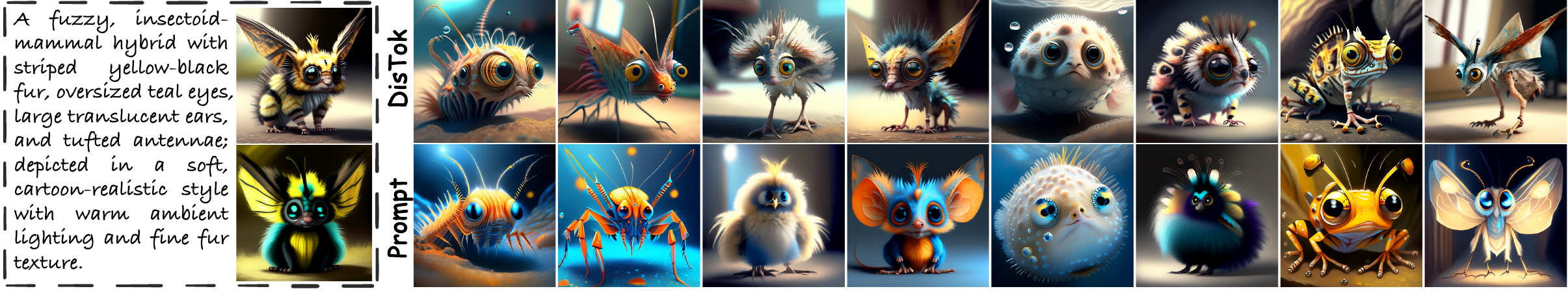}
  \vspace{-0.27in}
  \caption{Comparison with Prompt Engineering. For each DisTok-generated concept, a corresponding textual prompt for Kandinsky~\cite{razzhigaev2023kandinsky} is derived using GPT-4o~\cite{openai2023gpt4}. See Appendix~\ref{app:prompt} for all prompts.}
  \label{fig:prompt}
  \vspace{-0.2in}
\end{figure*}
Prompt engineering guides diffusion models toward creative outputs via handcrafted natural language prompts~\cite{almeda2024prompting, oppenlaender2024prompting}. However, it presupposes that users have already conceptualized the desired idea, shifting the burden of creativity to humans and limiting machine intelligence~\cite{lecun2022path}.
Nevertheless, we still compare DisTok with prompt engineering to illustrate DisTok’s generative capacity beyond linguistic expression.

For each image generated by DisTok, we extract a detailed description using GPT-4o and reuse it as a prompt in standard diffusion models (Fig.~\ref{fig:prompt}). Despite their linguistic richness, these prompts often yield images with compositional artifacts, or diminished semantic coherence—underscoring representational limits of natural language. 
In contrast, DisTok directly generates semantically rich, linguistically inexpressible concepts without human intervention.

\section{Conclusion}
Inspired by the tendency of classifiers to produce soft probability distributions over known classes when presented with unfamiliar inputs, we propose Distribution-Conditional Generation, a novel formulation that models creativity as image synthesis conditioned on class distributions. 
To this end, we present DisTok, an encoder-decoder framework that maps distributions into tokens of creative concepts, enabling both controllable generation and open-ended exploration within a unified architecture. 
Extensive experiments show that DisTok achieves state-of-the-art performance in creative generation with enhanced aesthetics, originality and semantic coherence.

\section*{Acknowledgement}
We sincerely thank Zedong Zhang, Wenqian Li, Jianlu Shen, and Ruixiao Shi for their insightful discussions on this work. We also appreciate Freepik for contributing to the figure design. 
This research is supported by the Science and Technology Major Project of Jiangsu Province under Grant BG20240305, Key Program of Jiangsu Science Foundation under Grant BK20243012, National Natural Science Foundation of China under Grants U24A20324, 62125602, and 62306073, Natural Science Foundation of Jiangsu Province under Grant BK20230832, and the Xplorer Prize.

\bibliography{neurips_2025}
\bibliographystyle{plain}

\clearpage
\appendix
\section{Additional Results}
\subsection{Additional Results for Distribution-Conditional Generation}
\label{app:dcg}
\begin{figure*}[t]
  \centering
  \vspace{0.5in}
  \includegraphics[width=\linewidth]{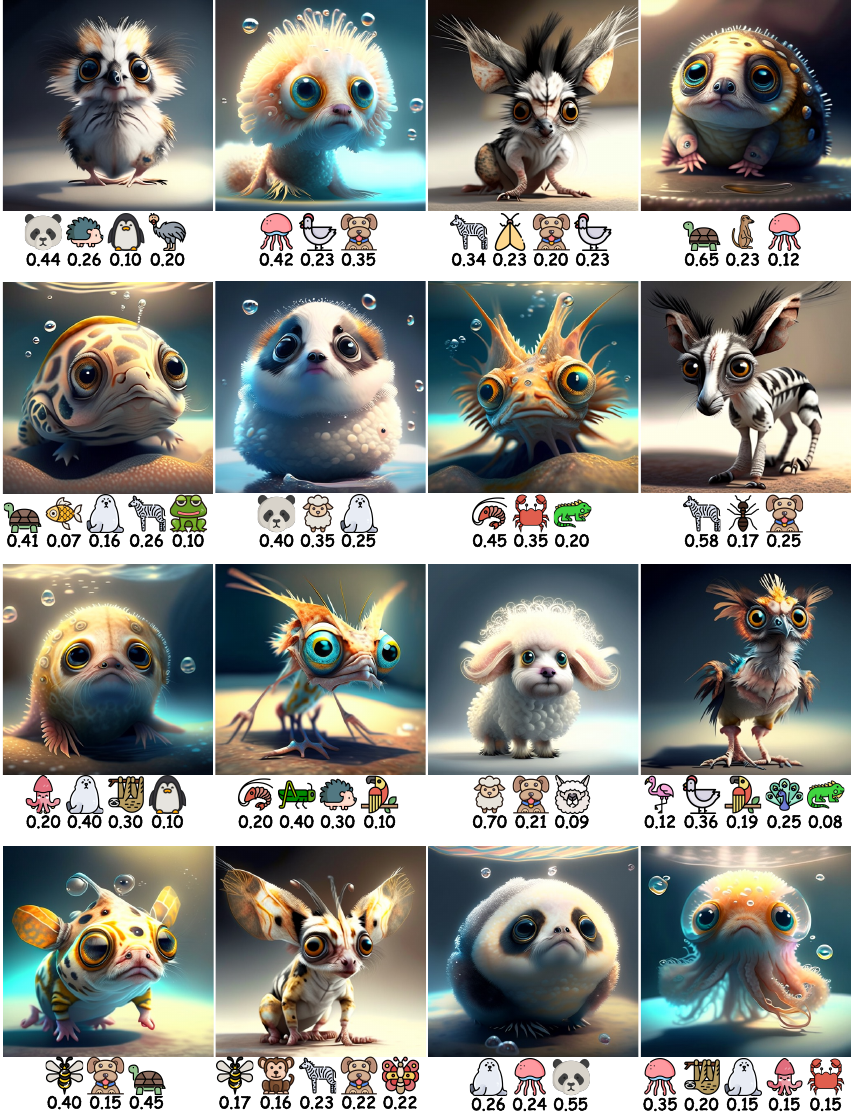}
  \vspace{-0.2in}
  \caption{Additional creative concepts generated by DisTok under Distribution-Conditional Generation task.}
  \vspace{0.5in}
  \label{fig:dcg_more}
\end{figure*}

We present additional qualitative results to further assess DisTok’s ability to perform fine-grained compositional generation under distribution-conditional generation settings. 
As illustrated in Fig.~\ref{fig:dcg_more}, DisTok effectively generates semantically coherent and visually integrated concepts across a broad range of input distributions, including those with more than three classes and highly imbalanced ratios.

Notably, DisTok preserves the target class proportions with high fidelity: dominant classes manifest through salient visual features, while minor classes contribute subtle but recognizable details. 
In contrast to baseline diffusion models, which often produce fragmented or spatially disjoint outputs (Fig.~\ref{fig:dcg}), DisTok achieves seamless fusion across concepts, even under complex distributional conditions. 
These results underscore the expressiveness of DisTok’s latent space and its robustness in modeling intricate, distribution-guided visual compositions.

\subsection{Additional Results for Text-pair to Object Task}
\label{app:image}
\begin{figure*}[t]
  \centering
  \includegraphics[width=\linewidth]{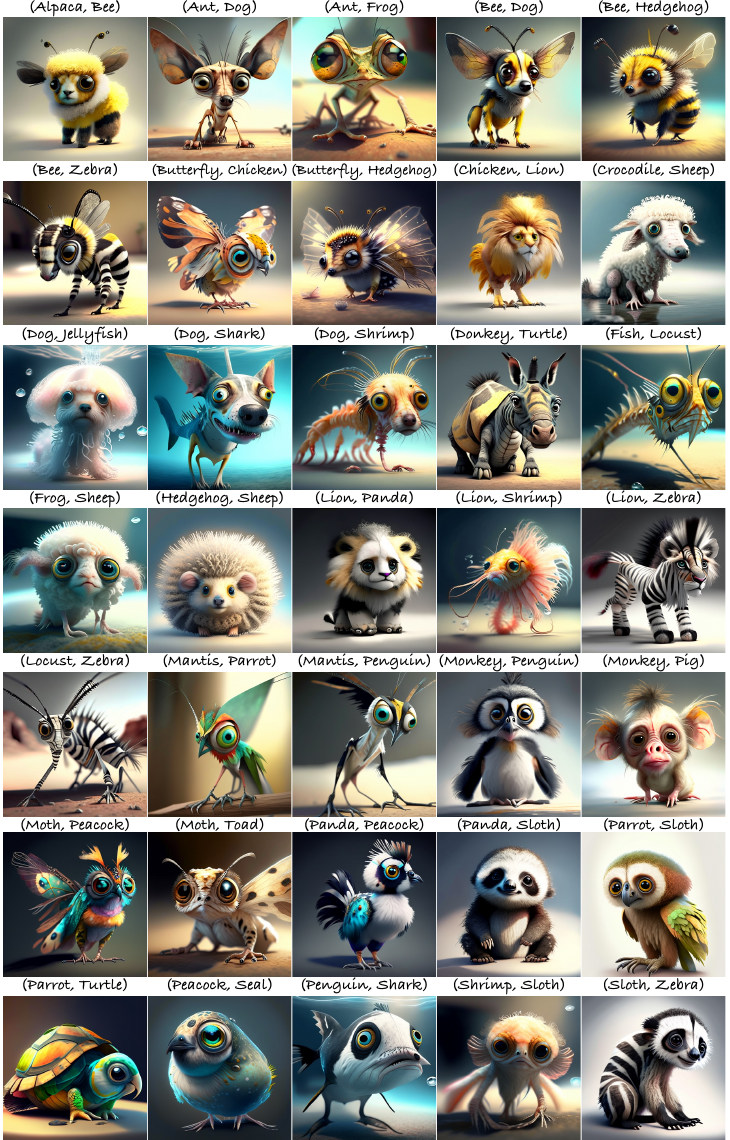}
  \vspace{-0.2in}
  \caption{Additional creative concepts synthesized by DisTok under the Text-Pair-to-Object (TP2O) task.}
  \label{fig:pair_more}
\end{figure*}

Fig.~\ref{fig:pair_more} presents additional qualitative results generated by DisTok under the Text-Pair-to-Object (TP2O) setting. Each example is conditioned on a concept pair, expressed either as two discrete classes or an equivalent class distribution (e.g., 50\% \texttt{Ant}, 50\% \texttt{Dog}).
These visualizations illustrate DisTok’s ability to synthesize coherent hybrid concepts that reflect integrated semantics rather than simple spatial composition or surface-level blending.

DisTok consistently produces unified and semantically coherent outputs under both balanced and imbalanced class distributions, preserving fine-grained attributes from each source concept. In comparison, token-based methods such as CreTok~\cite{feng2024redefining} and search-based approaches like BASS~\cite{litp2o} often suffer from significant computational overhead due to iterative sampling or optimization and struggle to maintain semantic balance—frequently resulting in dominant features or incomplete compositions (Fig.~\ref{fig:tp2o}). 
These shortcomings limit their scalability and applicability in practical settings. In contrast, DisTok enables efficient one-shot inference with strong compositional generalization, underscoring its effectiveness for controllable and scalable creative synthesis via distributional conditioning.

\subsection{Additional Results for Unconditional Open-Ended Creative Exploration}
\begin{figure*}[t]
  \centering
  \includegraphics[width=\linewidth]{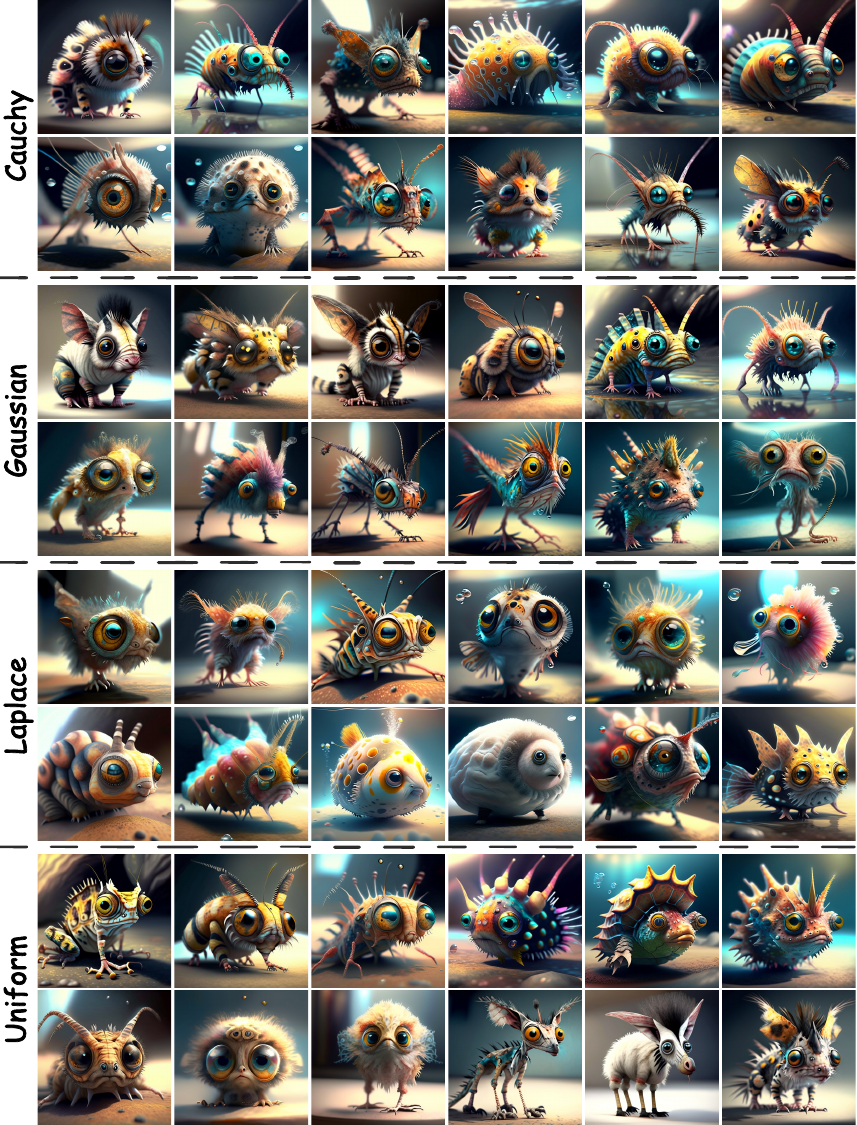}
  \vspace{-0.2in}
  \caption{Creative concepts generated unconditionally by DisTok through latent vector sampling from Cauchy, Gaussian, Laplace, and Uniform distributions.}
  \label{fig:latent_samples}
\end{figure*}

To further evaluate DisTok’s ability to support open-ended, unconditional generation, we present additional qualitative results obtained via direct latent sampling without any class-based conditioning. Latent vectors are sampled from multiple zero-mean, unit-variance distributions—including Gaussian, Cauchy, Laplace, and Uniform—and decoded into creative concept tokens. As shown in Fig.~\ref{fig:latent_samples}, DisTok produces diverse and semantically meaningful outputs across all distributions, highlighting the structure and generality of its latent space.

The generated concepts display substantial morphological and semantic diversity, demonstrating DisTok’s ability to synthesize novel visual forms beyond predefined prompts or control conditions. 
Unlike conditional generation paradigms grounded in text pairs or class distributions, which inherently constrain exploration to user-specified inputs, DisTok enables fully open-ended synthesis. 
These results highlight its potential as a scalable and autonomous framework for unconstrained concept discovery.

\subsection{Style-Aware Adaptability of Creative Concepts}
\label{app:style}
\begin{figure*}[t]
  \centering
  \includegraphics[width=\linewidth]{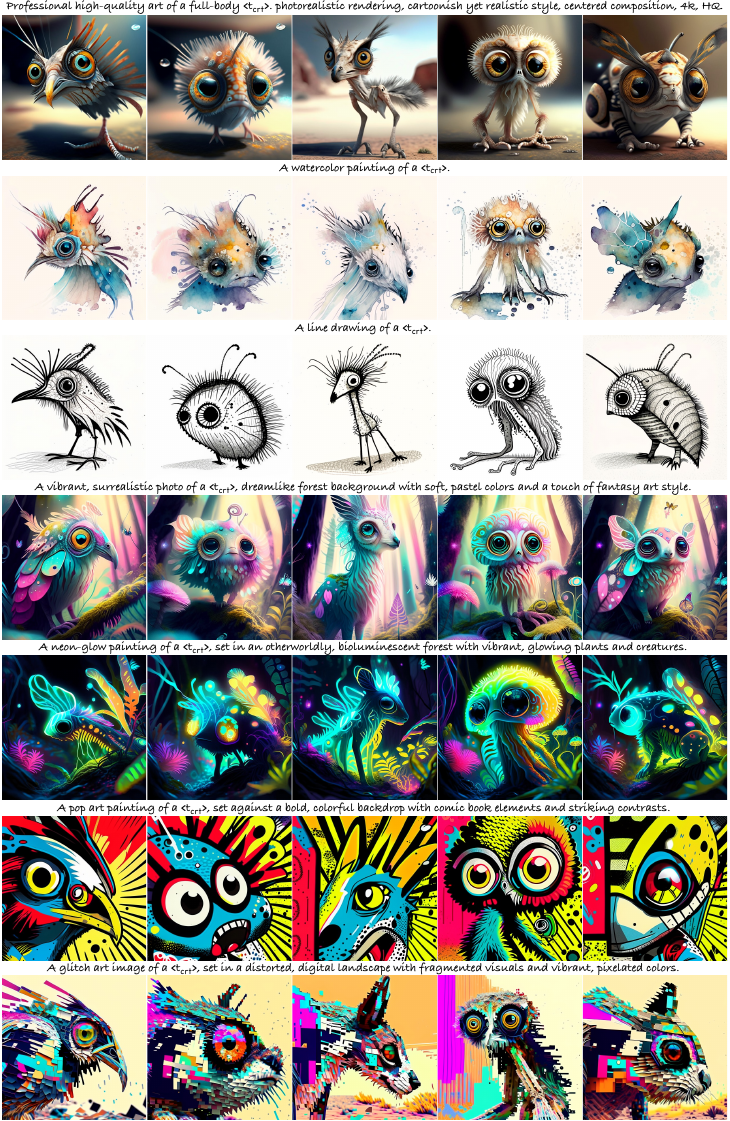}
  \vspace{-0.2in}
  \caption{Creative concepts rendered in diverse visual styles by integrating tokens generated by DisTok with style-specific natural language prompts.}
  \label{fig:app_style}
\end{figure*}

We assess the adaptability of DisTok-generated concepts to diverse visual styles by integrating creative tokens into natural language prompts that specify stylistic variations, such as watercolor painting, line drawing, and others (Fig.~\ref{fig:app_style}).

Across all styles, the generated images retain the semantic structure and visual identity of the original concepts, despite substantial changes in appearance induced by stylistic prompts. 
This consistency indicates that DisTok produces concept tokens that are semantically disentangled from style and can generalize across varied visual domains.
Such invariance to stylistic prompts highlights DisTok’s versatility in real-world creative workflows, including animation, game asset generation, and visual storytelling. 
It enables scalable and consistent concept deployment across heterogeneous visual settings, supporting controllable style adaptation while preserving conceptual fidelity.

\section{More Details on Evaluation}
\subsection{Prompts Used for GPT-4o Evaluation}
\label{app:prompt}
We conduct an objective evaluation of the creativity of images generated by DisTok and other methods using GPT-4o, assessing four key dimensions: Conceptual Integration, Alignment with Prompt, Originality, and Aesthetic Quality. 
The detailed evaluation prompts provided to GPT-4o are as follows, inspired by those proposed in CreTok~\cite{feng2024redefining}:

\textit{The subject of this evaluation is an image that represents a mixture of a Alpaca and a Bee (or a Zebra (34\%), a Moth (23\%), a Dog (20\%) and a Chicken (23\%)). The objective is to assess the creativity of an entity that synthesizes two distinct concepts as delineated in the provided prompt. Accordingly, please evaluate the creativity of images generated by various methodologies for the identical prompt, utilizing the following criteria on a scale from 1 to 10:}

\textit{1. Conceptual Integration (1-10): This criterion gauges the degree to which the image manifests a coherent and integrated concept, as opposed to merely placing two independent elements side by side. A high score signifies that the elements are intricately merged, creating a new, unified entity.}

\textit{2. Alignment with Prompt (1-10): This evaluates the extent to which the image conforms to and encapsulates the specific combination of concepts described in the prompt. The image should refrain from including irrelevant elements that detract from the primary concepts. A high score is allocated when the image closely adheres to the specifications of the prompt.}

\textit{3. Originality (1-10): This assesses the innovativeness of the concept portrayed in the image. The depicted concept should not mimic existing animals, plants, or widely recognized mythical creatures unless specifically mentioned in the prompt. Images that present a distinctive and inventive amalgamation receive a high score.}

\textit{4. Aesthetic Quality (1-10): This criterion scrutinizes the visual appeal of the image, focusing on color harmony, the balance and arrangement of elements, and the overall visual impact. A high score is awarded when the image is not only conceptually robust but also visually engaging.}

\textit{In conclusion, based on the aforementioned criteria, provide a comprehensive creative assessment (1-10) for each image and articulate specific justifications for your rating.}

\subsection{More Details on User Study}
\label{app:user}
\begin{wrapfigure}{r}{0.55\textwidth}
  \centering
  \vspace{-0.2in}
  \includegraphics[width=\linewidth]{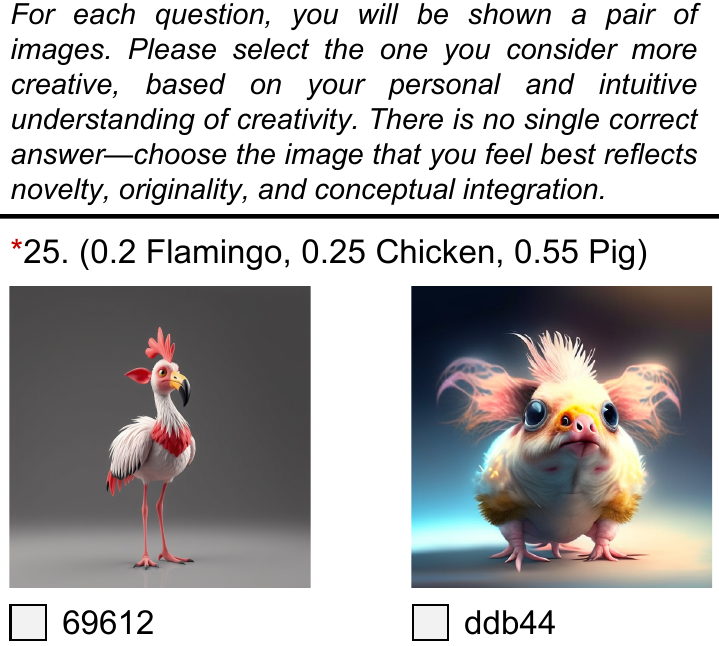}
  \vspace{-0.2in}
  \caption{Interface of the User Study.}
  \label{fig:app_infer}
\end{wrapfigure}

To ensure a comprehensive and balanced assessment of visual creativity, our user study includes 100 participants drawn from a diverse range of professional and non-professional backgrounds in art, design, and vision-related fields. This diversity allows us to evaluate whether the generated concepts satisfy both expert-level criteria and general audience preferences. The study interface is shown in Fig.~\ref{fig:app_infer}.

\clearpage
The participant composition is as follows:
\begin{itemize}
    \item \textbf{Art Researchers (10):} Scholars focused on art theory, history, and visual analysis, offering academically grounded evaluations of creativity.
    \item \textbf{Visual Arts Practitioners or Enthusiasts (15).} Individuals engaged in painting, sculpture, or digital art creation, with high sensitivity to form and composition.
    \item \textbf{Computer Vision / Graphics Researchers (17):} Researchers specializing in generative modeling, neural rendering, and visual synthesis, providing structurally informed assessments. 
    \item \textbf{Photographers and Photography Enthusiasts (13):} Professional photographers and dedicated hobbyists with experience in composition, lighting, and visual storytelling.
    \item \textbf{Design Professionals (18):} Practitioners in UI/UX, animation, architecture, and game design, with expertise in functional aesthetics and creative intent.
    \item \textbf{Art Educators (17):} Instructors in artistic disciplines with systematic training in evaluating visual creativity.
    \item \textbf{General Participants (10):} Individuals without formal creative backgrounds, included to reflect layperson judgments and aesthetic preferences.
\end{itemize}
This participant stratification ensures that the user study captures both domain-specific evaluations and intuitive, audience-centered perceptions of creativity.

\end{document}